\documentclass[10pt,twocolumn,letterpaper]{article}

\usepackage{cvpr}              %

\usepackage{graphicx}
\usepackage{amsmath}
\usepackage{amssymb}
\usepackage{booktabs}

\usepackage{bm}
\usepackage{multirow}
\usepackage[normalem]{ulem}
\useunder{\uline}{\ul}{}

\usepackage[pagebackref,breaklinks,colorlinks]{hyperref}

\usepackage[capitalize]{cleveref}
\crefname{section}{Sec.}{Secs.}
\Crefname{section}{Section}{Sections}
\Crefname{table}{Table}{Tables}
\crefname{table}{Tab.}{Tabs.}

\def\cvprPaperID{*****} %
\def\confName{UNCV-W}
\def\confYear{2023}

\begin{document}

\title{Low-Quality Image Detection by Hierarchical VAE}

\author{Tomoyasu Nanaumi, Kazuhiko Kawamoto, Hiroshi Kera\\
Chiba University, Japan \\
{\tt\small nanaumit428@chiba-u.jp, kawa@faculty.chiba-u.jp, kera@chiba-u.jp}
}
\maketitle

\begin{abstract}
To make an employee roster, photo album, or training dataset of generative models, one needs to collect high-quality images while dismissing low-quality ones.
This study addresses a new task of unsupervised detection of low-quality images.
We propose a method that not only detects low-quality images with various types of degradation but also provides visual clues of them based on an observation that partial reconstruction by hierarchical variational autoencoders fails for low-quality images.
The experiments show that our method outperforms several unsupervised out-of-distribution detection methods and also gives visual clues for low-quality images that help humans to recognize them even in thumbnail view.
\end{abstract}

\section{Introduction}
\label{sec:intro}
\begin{figure}[t] 
    \centering
    \includegraphics[width=\linewidth, bb=-10 30 855 510]{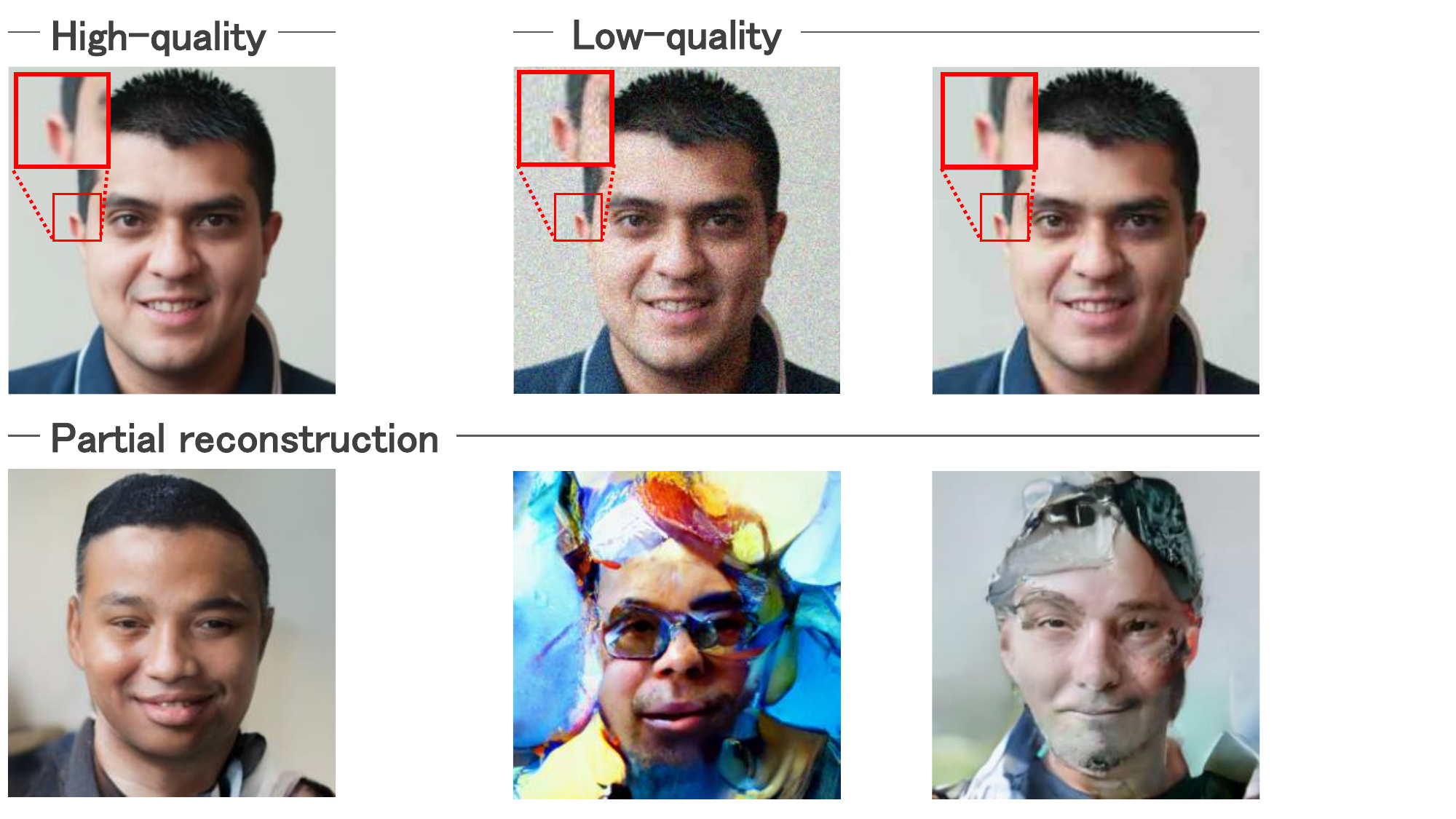} %
    \caption{Two low-quality variants with Gaussian noise and JPEG compression~(top center and top right) are difficult to detect for humans in thumbnail view. Partial reconstruction severely failed for these images~(bottom), serving as the base of the proposed method and as visual clues.}
    \label{fig:method}
  \end{figure}

When collecting many high-quality images, it is important to dismiss under-qualified ones in the collection process efficiently. Examples include making employee rosters, photo albums, or datasets for training generative models.
When one has to handle a huge number of images,
such tasks need to classify images automatically according to their quality and, if needed, manually for double check.
Training a machine-learning model to reject low-quality images is challenging because there are various sources of quality degradation, such as noise, blur, lighting, and digital processing~(e.g., JPEG compression). Even if one trains a model by synthesizing low-quality images using several corruptions, it is highly possible for this model not to perform well for unseen corruptions. Indeed, ~\cite{geirhos2018generalization} shows a poor generalization of multi-class classifiers trained on corrupted images to unseen corruptions.
Therefore, it is necessary to train a model that can detect diverse forms of image degradation without relying on synthesized low-quality images. 
Furthermore, it is also helpful for humans if the model provides a visual clue of low-quality images because the degradation is often unnoticeable in a thumbnail view~(in~\cref{fig:method}). 

In this study, we introduce a novel unsupervised binary classification task between high-quality and low-quality images.
Here, we consider common corruptions~\cite{Hendrycks_2019_ICLR_CommonCorruption} as a particular set of image degradation.
With our observation that corrupted images lead to a breakdown in \textit{partial reconstruction} by Very Deep VAE~\cite{Child_2021_ICLR_VDVAE}, 
we propose a detection method that rejects low-quality images based on the discrepancy between an original image and that from the partial reconstruction. 
The reconstructed images also serve as visual clues with which humans readily recognize corrupted images even in a gallery view as shown in~\cref{fig:method}.
The experiments show the effectiveness of the proposed method on low-quality image detection by comparing it with several unsupervised out-of-distribution~(OOD) detection methods~\cite{Havtorn_2021_ICML_LLRk, Tack_2020_NIPS_CSI, Reiss_2021_CVPR_PANDA}.
The proposed method achieves the best detection performance on the FFHQ-256 dataset~\cite{Karras_2019_CVPR_StyleGAN}, particularly for noises, blurs, JPEG compression, and saturation. 
The visualization shows that partial reconstruction gives significantly distorted outputs for corrupted images even with level-1 severity, which requires human eyes to enlarge the images to recognize them.

\section{Related Work}
\label{sec:related}
Our study addresses a new task of detecting low-quality images in an unsupervised manner. We here present several related detection tasks where certain types of noisy images and outliers are targeted.

The closest task to ours might be OOD detection,
which aims to detect data that do not belong to the training data distribution. Typically, only in-distribution samples are available for training~\cite{Ren_2019_NIPS_Likelihood_Ratio, Xiao_2020_NIPS_Likelihood_Regret, Serrà_2020_ICLR_IC, Havtorn_2021_ICML_LLRk, Li_2022_NIPS_LLRada, Chauhan_2022_CVPR_Robust, Bergman_2020_ICLR_GOAD, Qiu_2021_ICML_NeuTraL, Tack_2020_NIPS_CSI, Ruff_2018_ICML_DeepSVDD, Reiss_2021_CVPR_PANDA, Hendrycks_2019_ICLR_OE, Mirzaei_2023_ICLR_fake}.
Particularly, classification-based methods~\cite{Bergman_2020_ICLR_GOAD, Qiu_2021_ICML_NeuTraL, Tack_2020_NIPS_CSI, Ruff_2018_ICML_DeepSVDD, Reiss_2021_CVPR_PANDA, Hendrycks_2019_ICLR_OE, Mirzaei_2023_ICLR_fake} superior to generative model-based methods.
Our task is distinct from OOD detection, requiring a classification between clean and corrupted images, both in-distribution samples. Further, generative models are preferred as they provide visual clues.
Other related tasks include detection of images maliciously altered to fool models or humans, such as adversarial examples~\cite{Goodfellow_2015_ICLR_AdversarialExamples} and fake images~\cite{Karras_2019_CVPR_StyleGAN, Karras_2020_CVPR_StyleGAN2, Koujan_2020_FG_Head2Head, Liu_2022_ECCV_Semantic}. 
However, these tasks are usually tackled in supervised learning and the malicious images are available for training~\cite{Lu_2017_ICCV_SafetyNet, Kevin_2019_ICML_a_statistical, Yang_2020_AAAI_ML-LOO, Tramer_2022_ICML_Detecting, Agarwal_2022_CVPRW_Exploring, Zhao_2021_CVPR_Multi, Das_2021_ICCVW_Towards, Zhao_2021_ICCV_Learning, Chen_2022_CVPR_self, Guan_2022_NIPS_Delving}, whereas our study focuses on unsupervised learning to adapt to unseen types of degradation.

\section{Method}
\label{sec:method}
In this section, we first discuss our detection task and then present our key observation and the proposed method.

\begin{figure}[t]
    \centering
    \includegraphics[width=\linewidth, bb=20 230 970 520]{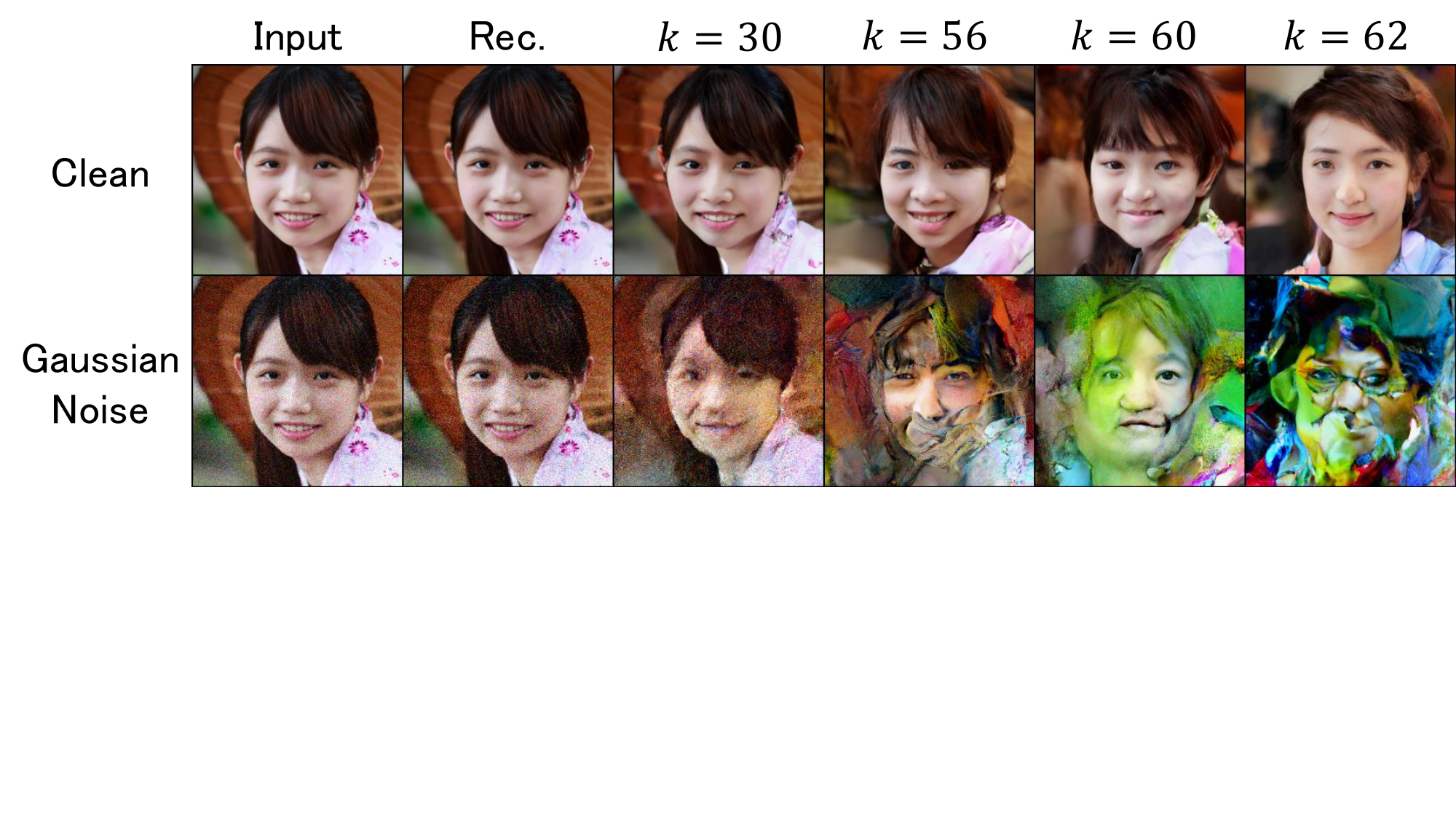}
    \caption{Comparison of input images, reconstructed images, and partial reconstructions for several $k$ for a clean image~(top row) and that with Gaussian noise~(bottom row). 
    }
    \label{fig:partial_generation_various_k}
\end{figure}
\subsection{Low-Quality Image Detection}
Given a collection of images $\mathcal{X} = \{\bm{x}_1, \ldots, \bm{x}_m\}$, our goal is the removal of low-quality images~(i.e., corrupted images) from $\mathcal{X}$. To this end, we want a score function $S(\bm{x})$ with the following properties: (i) it gives high scores for low-quality images, 
(ii) it generalizes well to various types of image degradation, and 
(iii) it gives a visual clue for humans to detect low-quality images readily.
We assume that only high-quality images are available for training.

Some of the motivating scenarios are as follows. 
For example, the Ministry of foreign affairs requires high-quality facial images for passport applicants. To handle a huge number of applications efficiently, a machine learning model can help it to reject low-quality ones automatically. 
Another example is the data collection for training generative models. Since such training requires a large-scale dataset, again, it is helpful if one can execute automatic data cleansing and remove low-quality images.
In both examples, a manual check is necessary to prevent accidental rejection and deletion.
However, noises and blurs may not be visible without zooming in, which makes the check time-consuming. Hence, we want to develop a method that detects low-quality images and visualizes the clues of those even in thumbnail view.

\subsection{Partial Reconstruction}
Hierarchical VAEs~\cite{Sonderby_2016_NIPS_HVAE} are an extension of the VAE~\cite{Kingma_2014_ICLR_VAE} framework. An $L$-layer model has hierarchical latent variables $\bm{z} = \{ \bm{z}_1, \bm{z}_2, \ldots, \bm{z}_L \}$ to model image distribution $p(\bm{x})$, where $\bm{z}_l$ denotes the latent variables of the $l$-th layer.
A hierarchical VAE consists of a generative model and an inference model, defined from the top-down architecture:
\begin{align}
 p_\theta (\bm{x} , \bm{z}) &= p_\theta (\bm{x} | \bm{z}_1) \prod_{l=1}^{L-1} p_\theta (\bm{z}_l | \bm{z}_{l+1}) p_\theta (\bm{z}_L), \\
 q_\phi (\bm{z} | \bm{x}) &= \prod_{l=1}^{L-1} q_\phi(\bm{z}_{l} | \bm{z}_{l+1}) q_\phi (\bm{z}_L | \bm{x}),
\end{align}
where $\theta, \phi$ denote the model parameters of the generative and inference models, respectively. The inference model is designed to approximate the posterior over latent variables.

The latent variables are partitioned into the higher-level set $\bm{z}_{>k} = \{\bm{z}_{k+1}, \ldots, \bm{z}_L\}$ and lower-level set $\bm{z}_{\leq k} = \bm{z} \setminus \bm{z}_{>k}$, where $k \in \{0, \ldots, L-1\}$. The partial generative model is then defined by $p_\theta (\bm{x} | \bm{z}_{>k})$. The image reconstruction using this model (and $q_\phi (\bm{z}_{>k} | \bm{x})$) is called partial reconstruction. In this paper, we also refer to the images reconstructed in this way as partial reconstructions when it is clear from the context.

At partial reconstruction, latent variables of layers after $k$ are sampled from the posterior distribution $q_\phi (\bm{z}_{>k} | \bm{x})$, while in the remaining $L-k$ layers, these are sampled from the prior distribution~(i.e., a normal distribution).
When $k=0$, partial reconstruction is equivalent to full reconstruction.
Several partial reconstructions with different $k$ are shown in~\cref{fig:partial_generation_various_k}.  As $k$ increases, information specific to the input images diminishes, and the partial reconstructions approach the mode of data distribution.
This means that the latent variables in higher layers of hierarchical VAEs affect common features of generated images, whereas the ones in lower layers affect more image-specific features.

In~\cref{fig:partial_generation_various_k}, our critical observation on the distinct outcome of partial reconstruction between clean and low-quality images is also shown.
The partial reconstructions for the image with Gaussian noise encounter a severe breakdown, whereas those for the clean image fall into the mode and are visually reasonable. This suggests that the detection of low-quality images can be achieved by observing the failure of partial reconstruction.

\begin{figure}[t]
    \centering
    \includegraphics[width=0.92\linewidth, bb=-20 0 440 510]{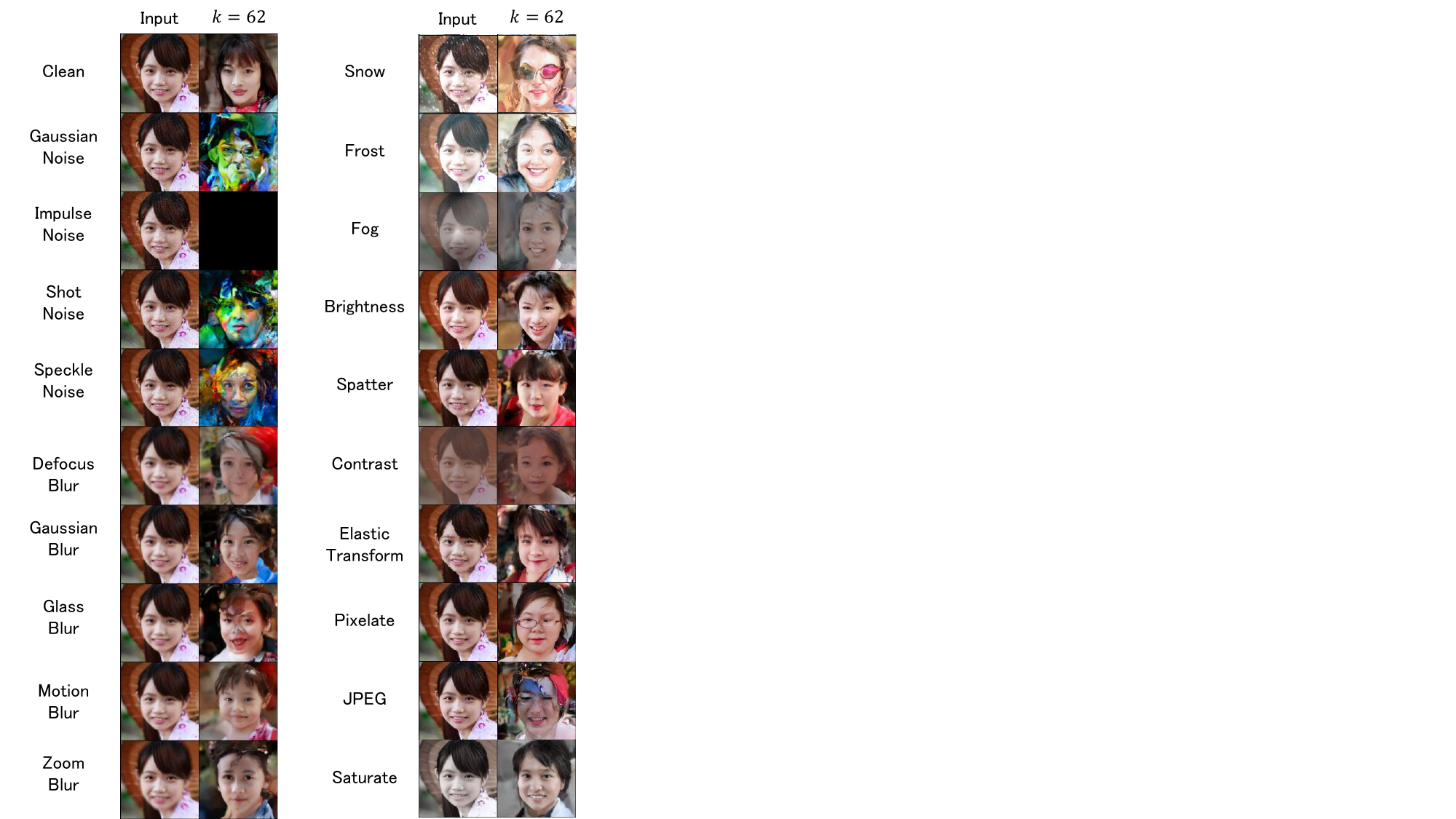}
    \caption{Comparison of inputs and partial reconstructions with each common corruption added as severity level 1 in VDVAE trained on FFHQ-256. The partial reconstructions are degraded by particularly noise and JPEG compression. For impulse noise, partial reconstruction fails, resulting in a black image.}
    \label{fig:partial_generation_c}
  \end{figure}

\subsection{Score Function}
To measure the failure of partial reconstruction, we now introduce a score function based on the dissimilarity of the original image and its partial reconstruction in the higher latent variables $\bm{z}_{>k}$. 
For clean images, their partial reconstructions tend to be similar in structure~(face direction, hair color, lighting, etc.). This implies a strong similarity between
the higher latent variables $\bm{z}_{>k}$ computed from a clean image and those from its partial reconstruction. 
In contrast, this should not be the case with low-quality images because, as \cref{fig:partial_generation_various_k} shows, such images lead to the breakdown in partial reconstruction.
 Based on this idea, we define the score function for input image $\bm{x}$ by  
 \begin{align}
     S_{\mathrm{KL}}(\bm{x}) = D_{\mathrm{KL}} \left[q_{\phi}(\bm{z}_{>k}|\bm x) \parallel q_{\phi}(\bm{z}_{>k}|\bm{x}_{\bm{z}_{>k}})\right], 
 \end{align}
 where $\bm{x}_{\bm{z}_{>k}}$ denotes the partial reconstruction of $\bm{x}$, and $D_{\mathrm{KL}}$ denotes the Kullback–Leibler~(KL) divergence.
 The score function measures the (pseudo-)distance between two latent probability distributions $q_{\phi}(\bm{z}_{>k}|\bm x)$ and $q_{\phi}(\bm{z}_{>k}|\bm{x}_{\bm{z}_{>k}})$ using the KL divergence.
 A high score suggests that the distribution of the latent variables of input image $\bm{x}$ are far apart from that of partial reconstruction $\bm{x}_{\bm z_{>k}}$, indicating that partial reconstruction fails and $\bm{x}$ is a degraded image.
 
To better measure the score, it may be desirable to adaptively select $k$ according to the characteristics of the degradation. We empirically observed that small $k$ is preferred for high-frequency degradation. Thus, we switch $k$ between $k_1, k_2$, where $k_1 < k_2$,  with the following steps: (i) calculate the frequency spectrum of an image by Fast Fourier Transform, 
(ii) calculate the mean $M$ of the higher-frequency spectrum by frequency filtering~(we set the filter size to the half of the image size), and 
(iii) use $k_1$ for partial reconstruction if $M > T$ with threshold $T$; otherwise, use $k_2$.
This procedure gives a better choice of $k$ for each image.

\begin{table*}[t]
\centering
\caption{Comparison of the proposed method and other unsupervised OOD detection methods for corrupted image detection on FFHQ-256. Each row shows the results with a dataset containing clean images (50~\%) and those with specific corruptions~(50~\%). The last row shows the average.
The best performance for each corruption is shown in bold. Arrows indicate the direction of better performance.}
\label{tab:result_FFHQ-256}
\scalebox{1.0}[1.0]{
\begin{tabular}{lrrrrrrrrrr}
\hline
\multicolumn{1}{c}{} & \multicolumn{5}{c}{AUROC↑}                                                  & \multicolumn{5}{c}{FPR80↓}                                           \\ \hline
 &
  \multicolumn{1}{l}{Likelihood} &
  \multicolumn{1}{l}{$\mathrm{LLR}^{>k}$} &
  \multicolumn{1}{l}{CSI} &
  \multicolumn{1}{l}{PANDA} &
  \multicolumn{1}{l}{Ours} &
  \multicolumn{1}{l}{Likelihood} &
  \multicolumn{1}{l}{$\mathrm{LLR}^{>k}$} &
  \multicolumn{1}{l}{CSI} &
  \multicolumn{1}{l}{PANDA} &
  \multicolumn{1}{l}{Ours} \\ \hline
Gaussian Noise       & \textbf{100} & 71.3         & 98.8          & 92.3          & \textbf{100}  & \textbf{0.0}  & 48.2 & 0.4           & 12.2          & \textbf{0.0}    \\
Impulse Noise        & \textbf{100} & \textbf{100} & 98.9          & 98.7          & \textbf{100}  & \textbf{0.0}  & 0.0  & 0.4           & 2.0           & \textbf{0.0}    \\
Shot Noise           & \textbf{100} & 75.4         & 99.3          & 92.7          & \textbf{100}  & \textbf{0.0}  & 40.0 & \textbf{0.0}  & 11.6          & \textbf{0.0}    \\
Speckle Noise        & \textbf{100} & 72.0           & 98.5          & 90.1          & 98.6          & \textbf{0.0}  & 46.7 & 1.0           & 14.4          & 2.2           \\
Defocus Blur         & 0.1          & 59.1         & 96.7          & 79.4          & \textbf{100}  & 100           & 65.1 & 2.4           & 31.7          & \textbf{0.0}    \\
Gaussian Blur        & 1.7          & 56.2         & 70.7          & 63.1          & \textbf{100}  & 99.9          & 70.0 & 52.6          & 58.0          & \textbf{0.0}    \\
Glass Blur           & 13.5         & 54.5         & 75.6          & 80.5          & \textbf{99.6} & 98.4          & 73.4 & 44.8          & 32.0          & \textbf{0.0}    \\
Motion Blur          & 6.2          & 50.9         & 75.0          & 61.5          & \textbf{99.6} & 99.7          & 77.6 & 45.2          & 61.3          & \textbf{0.0}    \\
Zoom Blur            & 1.3          & 46.9         & 82.1          & 74.0          & \textbf{100}  & 100           & 79.4 & 32.4          & 40.9          & \textbf{0.0}    \\
Snow                 & 94.1         & 67.4         & 96.2          & \textbf{98.7} & 97.8          & 7.1           & 51.4 & 2.8           & \textbf{1.9}  & 2.4           \\
Frost                & 90.8         & 66.8         & 92.5          & 89.8          & \textbf{93.4} & 13.1          & 55.8 & \textbf{11.0} & 17.5          & 15.0            \\
Fog                  & 3.3          & 16.8         & 86.1          & 76.3          & \textbf{99.9} & 99.9          & 93.1 & 24.4          & 37.7          & \textbf{1.0}    \\
Brightness           & 64.9         & 56.2         & \textbf{55.0} & 52.5          & 46.6          & \textbf{61.4} & 73.6 & 76.2          & 76.2          & 76.3          \\
Spatter              & 60.8         & 54.3         & 55.6          & \textbf{69.2} & 67.5          & 65.5          & 73.0 & 75.8          & \textbf{51.0} & 58.2          \\
Contrast             & 2.6          & 8.7          & 65.9          & 70.3          & \textbf{99.8} & 99.9          & 98.3 & 60.8          & 48.3          & \textbf{0.0}    \\
Elastic Transform    & 43.4         & 50.1         & 65.9          & \textbf{88.5} & 80.0            & 84.4          & 78.6 & 60.4          & \textbf{19.5} & 44.1          \\
Pixelate             & 72.6         & 56.6         & 56.4          & 72.3          & \textbf{77.4} & 49.1          & 73.7 & 74.6          & 44.2          & \textbf{38.2} \\
JPEG Compression     & 56.4         & 60.4         & 68.5          & 86.0          & \textbf{97.4} & 69.4          & 65.1 & 55.2          & 22.7          & \textbf{2.3}  \\
Saturate             & 28.4         & 47.5         & 60.2          & 74.7          & \textbf{77.6} & 93.4          & 76.5 & 70.2          & 38.9          & \textbf{34.7} \\ \hline
Average              & 49.5         & 56.4         & 78.8          & 79.5          & \textbf{91.3} & 60.1          & 65.2 & 36.3          & 32.7          & \textbf{14.4} \\ \hline
\end{tabular}
}
\end{table*}

\begin{table}[t]
\centering
\caption{The detection performance of the proposed method and other unsupervised OOD detection methods. Each of FFHQ-256~(FFHQ) and ImageNet-64~(IN) consist of clean images (50~\%) and those with common corruptions (50~\%). All 19 types of corruption are included.}
\label{tab:result_average}
\scalebox{0.98}[0.98]{
\begin{tabular}{llrrr}
\hline
Dataset & \multicolumn{1}{l}{Method} & \multicolumn{1}{c}{AUROC↑} & \multicolumn{1}{c}{AUPRC↑} & \multicolumn{1}{c}{FPR80↓} \\ \hline
\multirow{6}{*}{FFHQ} & Likelihood & 40.6          & 60.6          & 99.6          \\
                      & $\mathrm{LLR}^{>k}$        & 54.2          & 51.0          & 77.4          \\
                      & CSI        & 78.8          & 75.9          & 36.6          \\
                      & PANDA      & 79.7          & 81.9          & 38.1          \\
                      & Ours~($k=54$) & 80.9          & 81.6          & 34.4          \\
                      & Ours       & \textbf{90.0} & \textbf{92.5} & \textbf{10.0} \\ \hline
\multirow{6}{*}{IN}   & Likelihood & 66.8          & 73.9          & 74.5          \\
                      & $\mathrm{LLR}^{>k}$        & 37.6          & 43.5          & 95.7          \\
                      & CSI        & 69.4          & 68.4          & 52.6          \\
                      & PANDA      & \textbf{79.9} & \textbf{82.2} & \textbf{37.9} \\
                      & Ours~($k=35$) & 74.4          & 77.1          & 51.6          \\
                      & Ours       & 76.2          & 77.4          & 46.6          \\ \hline
\end{tabular}
}
\end{table}

\section{Experiments}
\label{sec:experiments}
We evaluate our method on FFHQ-256~\cite{Karras_2019_CVPR_StyleGAN} and ImageNet-64~\cite{ImageNet-64} in the low-quality image detection task. 
For each of the two datasets, we added 19 types of common corruptions using the official package of~\cite{Michaelis_2019_ImageCorruptions}. This package was used for constructing corrupted datasets such as CIFAR-10-C~\cite{Hendrycks_2019_CIFAR10C} and ImageNet-C~\cite{Hendrycks_2019_ICLR_CommonCorruption}. The severity of common corruption is set to level 1, the hardest setting for classifying corrupted images from clean ones. 
For partial reconstruction, we used Very Deep VAEs\footnote{\url{https://github.com/openai/vdvae}}~(VDVAEs~\cite{Child_2021_ICLR_VDVAE}) pretrained on FFHQ-256 and ImageNet-64 with the number of layers $L=66, 75$, respectively.
For the frequency-based choice of $k$, we used $(k_1, k_2, T) = (36, 54, 1.8), (25, 55, 4.3)$ for FFHQ-256 and ImageNet-64, respectively.
We adopted several OOD detection methods as baselines: 
(i) VDVAE with likelihood score function, 
(ii) VDVAE with $\mathrm{LLR}^{>k}$ score function~\cite{Havtorn_2021_ICML_LLRk}, 
(iii) CSI\footnote{\url{https://github.com/alinlab/CSI}}~\cite{Tack_2020_NIPS_CSI}, and 
(iv) PANDA\footnote{\url{https://github.com/talreiss/PANDA}}~\cite{Reiss_2021_CVPR_PANDA}. 
CSI and PANDA are two of the best OOD detection methods that do not require additional data such as outlier exposure~\cite{Hendrycks_2019_ICLR_OE}.
We used three standard evaluation metrics, AUROC~(Area Under the Receiver Operating Characteristic curve), AUPRC~(Area Under the Precision-Recall Curve), and FPR80~(False Positive Rate at 80~\% True Positive Rate), for a binary classification task on clean test images and those with common corruptions.

Figure~\ref{fig:partial_generation_c} shows a clean image and those with common corruptions on the left of each column and their partial reconstructions on the right.
It is hard for human eyes to recognize degradation in thumbnail size, especially in the case of noises, blurs, and JPEG compression. 
In contrast, the partial reconstructions of the corrupted images, especially noises, Glass blur, Snow, or JPEG compression, show significant distortions in facial color and structure. These serve as visual clues that help humans notice the presence or absence of corruption, even in thumbnail display.

Table~\ref{tab:result_FFHQ-256} shows AUROC and FPR80 of the detection methods on FFHQ-256.
The proposed method achieved the highest detection accuracy for various cases and the average. It performs particularly well for noises, blurs, and JPEG compression.
Among VAE-based methods, our score function detects corrupted images with stable accuracy across corruption types, while the likelihood and $\mathrm{LLR}^{>k}$ perform unstably. 
Table~\ref{tab:result_average} shows the detection performance on the datasets containing clean images and all types of corruptions at an even ratio.
The proposed method achieved the best performance on FFHQ-256. It performs slightly worse than PANDA on ImageNet-64 but far better than other methods, including CSI.
Furthermore, the frequency-based choice of $k$ leads to higher performance than fixed $k$.
Note that in the literature, VAE-based OOD detection methods perform significantly worse than classification-based methods such as PANDA and CSI. This study introduced a task where VAE-based methods are potentially advantageous over classification-based methods in detection and visual clues.

\section{Conclusion} 
\label{sec:conclusion}
In this study, we tackled a novel problem of detecting low-quality images.
Based on our critical observation that corrupted images lead to a significant failure of partial reconstruction by hierarchical VAE, we proposed a low-quality image detection method using partial reconstruction.
The experiments demonstrate that the proposed method provides helpful visual clues for manual checks and 
significantly outperforms most unsupervised OOD detection methods on FFHQ-256 and ImageNet-64. 

\section*{Acknowledgements}
\noindent This work was supported by JSPS KAKENHI Grant Number JP22H03658, JP22K17962.

 \newcommand{\noop}[1]{}

\end{document}